\documentclass[journal]{IEEEtran}
\ifCLASSINFOpdf
  \usepackage[pdftex]{graphicx}
\else
\fi
\usepackage{amsmath}

\usepackage{multirow}
\usepackage{booktabs}

\begin{document}
%
\title{Look Wider to Match Image Patches with Convolutional Neural Networks}
%
%
%
\author{Haesol~Park,
and~Kyoung~Mu~Lee
\thanks{H. Park and K. M. Lee are with Automation and Systems Research Institute, Seoul National University, Seoul 151-744, Korea}}

\markboth{IEEE Signal Processing Letters}%
{H. Park \MakeLowercase{\textit{et al.}}: Look Wider to Match Image Patches with CNN}
%



\maketitle

\begin{abstract}
When a human matches two images, the viewer has a natural tendency to view the wide area around the target pixel to obtain clues of right correspondence.
However, designing a matching cost function that works on a large window in the same way is difficult.
The cost function is typically not intelligent enough to discard the information irrelevant to the target pixel, resulting in undesirable artifacts.
In this paper, we propose a novel convolutional neural network (CNN) module to learn a stereo matching cost with a large-sized window.
Unlike conventional pooling layers with strides, the proposed per-pixel pyramid-pooling layer can cover a large area without a loss of resolution and detail.
Therefore, the learned matching cost function can successfully utilize the information from a large area without introducing the fattening effect.
The proposed method is robust despite the presence of weak textures, depth discontinuity, illumination, and exposure difference. The proposed method achieves near-peak performance on the Middlebury benchmark.
\end{abstract}


%
\IEEEpeerreviewmaketitle

\section{Introduction}
Most stereo matching methods first compute the matching cost of each pixel with a certain disparity, before optimizing the whole cost volume either globally or locally by using specific prior knowledge~\cite{scharstein2002taxonomy}.
For decades, many researchers have focused on the second step, designing a good prior function and optimizing it~\cite{kolmogorov2001computing,hirschmuller2008stereo,woodford2009global,rhemann2011fast,yang2012non}.
Few studies have been conducted on designing or selecting a better matching cost function.

\begin{figure}[t!]
\centering
	\includegraphics[width=0.49\textwidth]{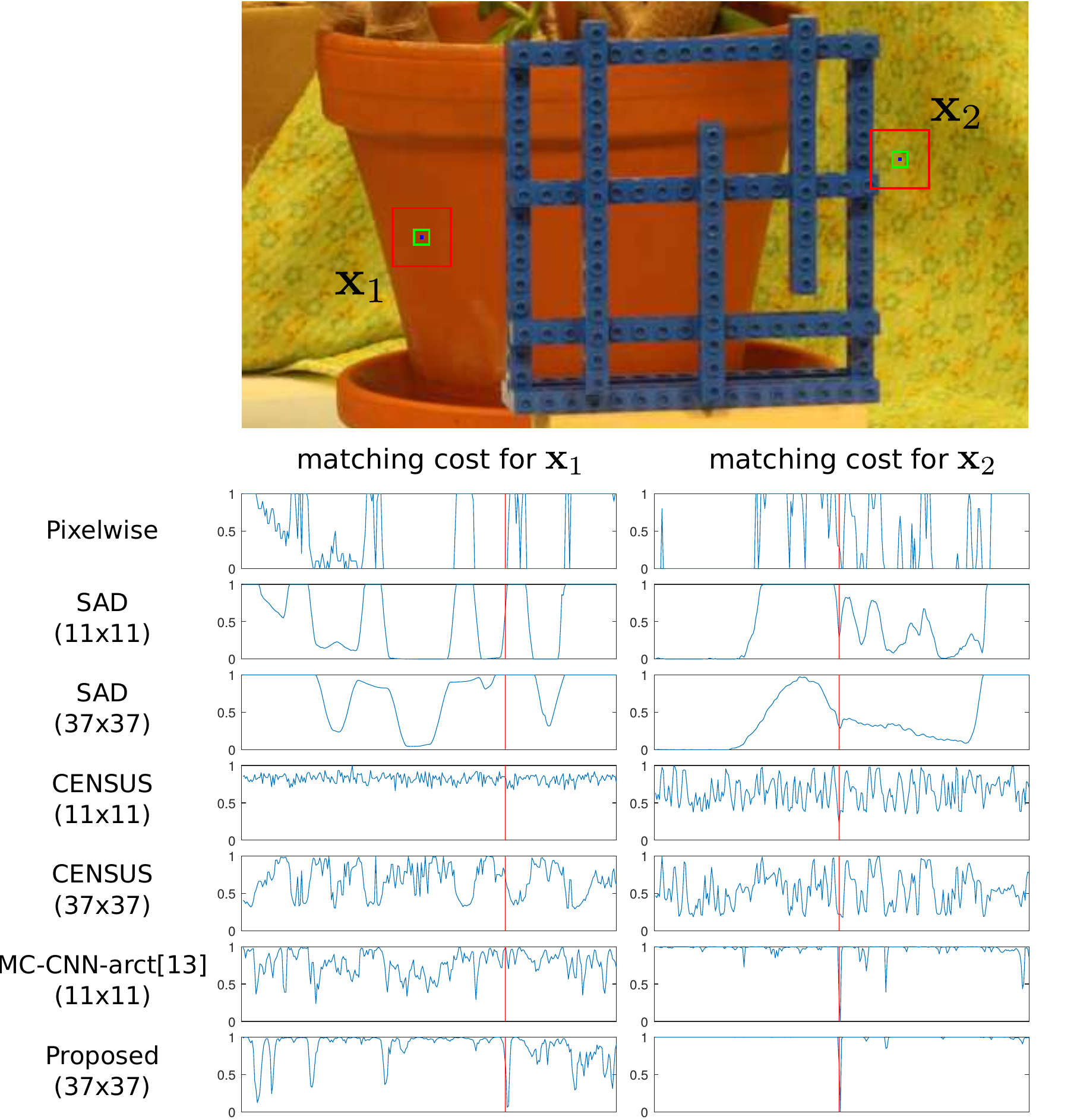}
	\caption{The top image shows the reference image with  two interested points, $x_1$ and $x_2$. The pixel positions are marked as  blue dots, whereas the red and green boxes represent $37\times37$  and $11\times11$ windows centered on them, respectively. At the bottom, the matching costs for each pixel are visualized as a normalized function of disparity for different matching cost functions. The positions of true disparities are marked as red vertical lines. The pixel-wise cost shows the lowest values at the true disparity, but it also gives zero costs for other disparities. The SAD and CENSUS matching cost functions~\cite{hirschmuller2007evaluation} become less ambiguous as the matching window becomes larger. However, these functions are affected by pixels irrelevant to the target pixel (${{\bf{x}}_2}$). Even the matching cost learned by using the baseline convolutional neural network (CNN) architecture fails when the surface has a nearly flat texture (${{\bf{x}}_1}$). On the other hand, the proposed CNN architecture works well both on weakly textured regions and disparity discontinuities.}
\label{fig:intro}
\end{figure}

One of the most widely used matching cost functions is a pixel-wise matching cost function, such as the one used in~\cite{birchfield1999depth}.
Along with sophisticated prior models, it sometimes produces good results, especially in preserving the detailed structures  near the disparity discontinuities.
However, the function  fails when the image contains weakly-textured areas or repetitive textures.
In such cases, a window-based matching cost, such as \textbf{CENSUS} or \textbf{SAD}~\cite{hirschmuller2009evaluation}, produces a more reliable and distinctive measurement.
The critical shortcoming of window-based matching cost functions is their unreliability around disparity discontinuities.
Figure~\ref{fig:intro} visually illustrates the characteristics of different matching cost measures.

One method to handle this trade-off is to make the window-based
versatile to its input patterns~\cite{wang2004adaptive,yoon2006adaptive,tombari2008classification}. The key idea is making the shape of the matching template adaptive so that it can discard the information from the pixels that are irrelevant to the target pixel. However, knowing the background pixels before the actual matching is difficult, making it a \lq{}chicken-and-egg\rq{} problem.
Therefore, the use of a CNN~\cite{vzbontar2016stereo,ZagoruykoCVPR2015} is appropriate, as it automatically learns the proper shape of the templates for each input pattern.

The existing methods, however, are based on conventional CNN architectures resembling the \textbf{AlexNet}~\cite{krizhevsky2012imagenet} or \textbf{VGG}~\cite{simonyan2014very} network, which are optimized for image classification task and not for image matching.
The architectures comprise several convolution layers, each followed by a rectified linear unit~(ReLU)~\cite{krizhevsky2012imagenet}, and pooling layers with strides.
One of the limitations of using these architectures for matching is  the difficulty of enlarging the size of the patches that are to be compared.
The effective size of the patch is directly related to the spatial extent of the receptive field of CNN, which can be increased by (1) including a few strided pooling/convolution layers, (2) using larger convolution kernels at each layer, or (3) increasing the number of layers.
However, use of strided pooling/convolution layers makes the results downsampled, losing fine details.
Although the resolution can be recovered by applying fractional-strided convolution~\cite{radford2015unsupervised}, reconstructing small or thin structures is still difficult if once they are lost after downsampling.
Increasing the size of the kernels is also problematic, as the number of feature maps required to represent the larger patterns increases significantly.
Furthermore, a previous study~\cite{zhou2014object} reported that the repetitive usage of small convolutions does not always result in a large receptive field.


This paper contributes to the literature by proposing a novel CNN module to learn a better matching cost function.
The module is an innovative pooling scheme that enables a CNN to view a larger area without losing the fine details and without increasing the computational complexity during test times.
The experiments show that the use of the proposed module improves the performance of the baseline network, showing competitive results on the Middlebury~\cite{scharstein2002taxonomy,scharstein2014high} benchmark.

\section{Related Works}
Given the introduction of high-resolution stereo datasets with the ground-truth disparity maps~\cite{Geiger2012CVPR,scharstein2014high,Menze2015CVPR}, many attempts have been made to learn a matching cost function using machine learning algorithms~\cite{vzbontar2016stereo,ZagoruykoCVPR2015,ladicky2015learning}.
The most impressive results are obtained by using CNN~\cite{vzbontar2016stereo,ZagoruykoCVPR2015}.
The architecture proposed in ~\cite{vzbontar2016stereo} takes a small $11\times11$ window and processes it without the use of pooling.
The computed cost volume is noisy due to the limited size of the window.
Thus, it is  post-processed by using the cross-based cost aggregation (CBCA)~\cite{zhang2009cross}, semi-global matching (SGM)~\cite{hirschmuller2008stereo}, and additional refinement procedures.
On the other hand, the method in~\cite{ZagoruykoCVPR2015} uses multiple pooling layers and spatial-pyramid-pooling (SPP)~\cite{he2014spatial} to process larger patches. However, the results show a fattening effect owing to the loss of information introduced by pooling.

The main contribution of this paper is in proposing a novel pooling scheme that can handle information from a large receptive field without losing the fine details.
Recently, several attempts have been made to accomplish the same goal in the context of semantic segmentation~\cite{long2015fully,hariharan2015hypercolumns,noh2015learning}.
These methods combine the feature maps from the high-level layers with those from the lower layers, with the aim of correctly aligning the object-level information along the pixel-level details.
While this approach can successfully align the boundaries of the big objects, its inherent limitation is its inability to recover small objects in the final output once they are lost during the abstraction due to multiple uses of pooling.
In the same context, the \emph{FlowNet}~\cite{fischer2015flownet} architecture can upsample the coarse-level flow to the original scale by using lower-level feature maps.
However, it fails to recover the extreme flow elements that are hidden due to the low resolution of high-level feature maps.

The architecture most closely related to the current work has been proposed in~\cite{he2014spatial}. Unlike the other approaches, the SPP network excludes pooling layers between convolutional layers.
Instead, it first computes highly-nonlinear feature maps by cascading convolutional layers several times and then generates high-level and mid-level information by pooling them at different scales.
By keeping the original feature maps along with feature maps pooled at multiple scales, the SPP network can combine the features from multiple levels without losing fine details.
Although the previously mentioned stereo method in~\cite{ZagoruykoCVPR2015} uses SPP, it also employs conventional pooling layers between convolutional layers, thus losing the detailed information.

\section{Architecture of the Neural Network}
The proposed architecture takes two input patches and produces the corresponding matching cost.
In the following subsections, the newly proposed module is first introduced. Then the detailed architecture of the entire network is presented.

\begin{figure}[t]
\centering
	\includegraphics[width=0.4\textwidth]{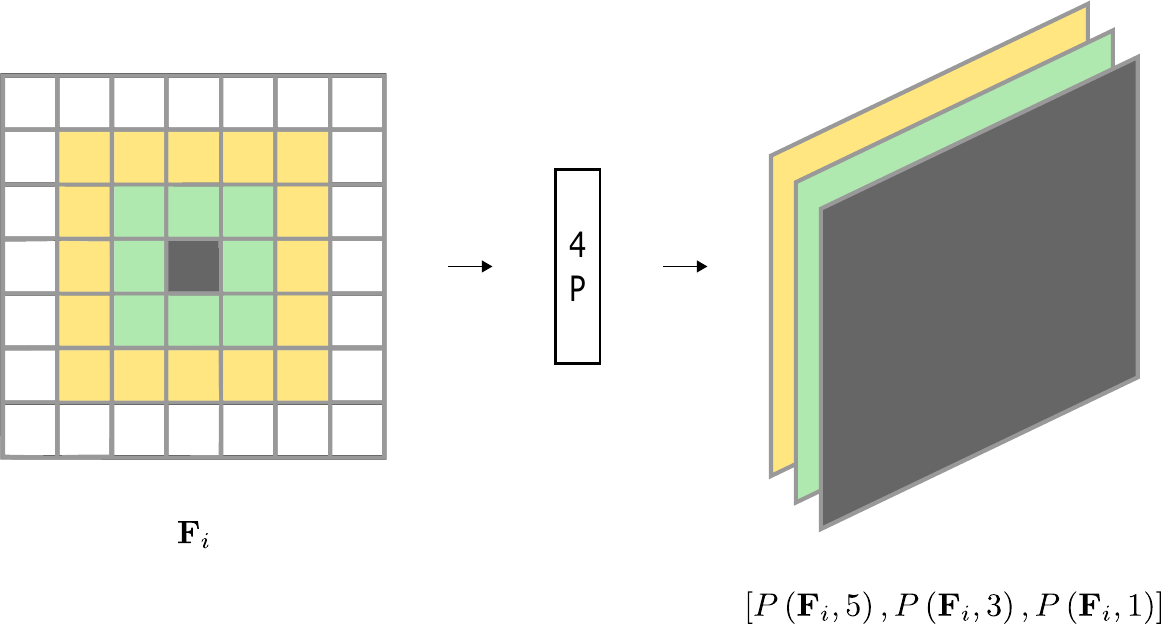}
	\caption{The \textbf{4P} module with pooling size vector ${\bf{s}} = \left[ {5,3,1} \right]$ is visualized. This figure shows its action for one channel of the feature maps for brevity; it does the same job for all channels.}
\label{fig:4Pmodule}
\end{figure}

\subsection{Per-pixel Pyramid Pooling (\textbf{4P})}
The use of pooling layers in CNN has been considered desirable because of its accuracy and efficiency in image classification tasks.
While the use of max-pooling layers has been reported to provide an additional invariance in spatial transformation, the most important gain comes from the downsampling of feature maps.
By performing pooling with a stride that is larger than one, the output feature maps after the pooling are scaled down.
The final scale of the CNN output is decreased exponentially in terms of the number of pooling layers.
Given that no parameters related to a pooling operation exist, this method is an effective way to widen the receptive field area of a CNN without increasing the number of parameters.
The drawback of strided pooling is that the network loses fine details in the original feature maps as the pooling is applied.
Thus, a trade-off exists in seeing a larger area and preserving the small details.

Inspired by the idea discussed in~\cite{he2014spatial}, we propose a novel pooling scheme to overcome this trade-off. Instead of using a small pooling window with a stride, a large pooling window is used to achieve the desired size of the receptive field. The use of one large pooling window can lead to the loss of finer details.
Thus, multiple poolings with varying window sizes are performed, and the outputs are concatenated to create new feature maps. The resulting feature maps contain the information from coarse-to-fine scales. The multi-scale pooling operation is performed for every pixel without strides.

We call this whole procedure, \lq\lq{}per-pixel pyramid pooling\rq\rq{} (\textbf{4P}), which is formally defined as follows:
\begin{equation}
\label{eq:4p}
{P^{4P}}({\bf{F}},{\bf{s}}) = \left[ {P\left( {{\bf{F}},{{s}_1}} \right), \cdots ,P\left( {{\bf{F}},{{s}_M}} \right)} \right],
\end{equation}
where ${\bf{s}}$ is a vector having $M$ number of elements, and $P\left( {{\bf{F}},{s_i}} \right)$ is the pooling operation with size $s_i$ and stride one. The structure of this module is illustrated in Figure~\ref{fig:4Pmodule}.

\subsection{Proposed model}
\label{sec:proposedModel}
\begin{figure}[t!]
\centering
	\includegraphics[width=0.45\textwidth]{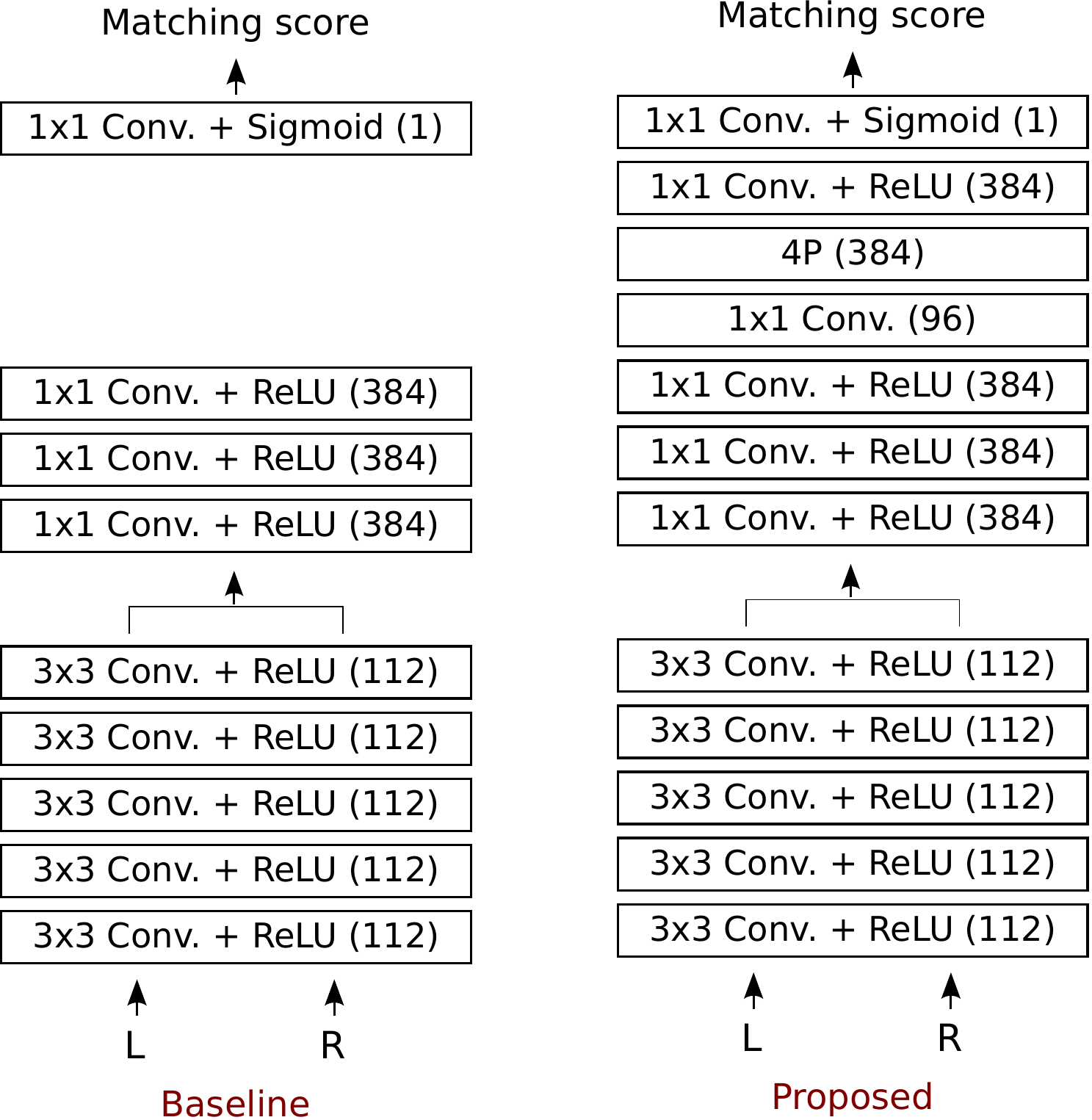}
	\caption{The network structures are visualized for the baseline network, \lq{}\textbf{MC-CNN-acrt}\rq{}~\cite{vzbontar2016stereo}, and the proposed network. The parenthesized numbers at each layer represent the number of feature maps after the corresponding operations. Note that this figure is drawn in terms of the fully convolutional network.}
\label{fig:CNN}
\end{figure}

To validate the effect of the proposed module, we trained and tested CNNs with and without the \textbf{4P} module.
The baseline architecture is selected as the \lq{}\textbf{MC-CNN-acrt}\rq{}~\cite{vzbontar2016stereo}.
The \textbf{4P} module in the proposed architecture is constructed by using the size vector ${\bf{s}} = \left[ {27, 9, 3, 1} \right]$.
The structures of two CNNs are visualized in Figure~\ref{fig:CNN}.

%

\section{Implementation Details}
For a fair comparison, we followed the details in~\cite{vzbontar2016stereo} to train the proposed architecture with a few exceptions mentioned below.
First, the size of the training patch became $37\times37$.
Furthermore,  we only fine-tuned the parameters of the last three $1\times1$ convolution layers of the proposed architecture in Figure~\ref{fig:CNN}.
The parameters of the earlier layers are borrowed from the pre-trained \lq{}\textbf{MC-CNN-acrt}\rq{}~\cite{vzbontar2016stereo} network.
In our experiments, this resulted in a better performance than the end-to-end training of the network with random initializations.
Moreover, training a few convolution layers with pre-trained features is easier, making it  converge faster.
We have run a total of four epochs of training, where the last two epochs were run with a decreased learning rate from 0.003 to 0.0003.


We also used the same post-processing pipeline as in~\cite{vzbontar2016stereo} during the test phase.
The post-processing pipeline includes the use of the CBCA~\cite{zhang2009cross} and SGM~\cite{hirschmuller2008stereo}, and the disparity maps are refined to have continuous values and undergo median filtering and bilateral filtering.


\section{Experiments}
\label{sec:experiments}
To verify the effect of the proposed  \textbf{4P} module, we have compared the results of the baseline and proposed network.
The performance is measured using the \lq{}training dense\rq{} set of the Middlebury benchmark~\cite{scharstein2002taxonomy}.
The quantitative results are briefly summarized in Table~\ref{tab:quan_exp1} using the average errors. All experiments are performed by using the Intel core i7 4790K CPU and a single Nvidia Geforce GTX Titan X GPU.

\begin{table}[t]
\caption{The quantitative results on the  \lq{}training dense\rq{} set of the Middlebury benchmark~\cite{scharstein2002taxonomy} are shown. The error represents the percentage of bad pixels with a disparity threshold 2.0, and the same weighting scheme is  applied as in~\cite{scharstein2002taxonomy}  when computing the average. }
\centering
\begin{tabular}{c l r}
\toprule
 \multicolumn{2}{c}{\textbf{Methods}} & \textbf{avg. error} \\
\midrule
\multirow{2}{*}{WTA} & \textbf{MC-CNN-acrt~\cite{vzbontar2016stereo}} & 22.91\\
& proposed & 11.75\\
\midrule
\multirow{4}{*}{after post-processing} & \textbf{MC-CNN-acrt~\cite{vzbontar2016stereo}} & 10.26 \\
& proposed  & \multirow{2}{*}{9.72}\\
& (w/ parameters in~\cite{vzbontar2016stereo}) \\
& proposed  & \multirow{2}{*}{8.45}\\
& (w/ parameter tuning) \\
\bottomrule
\end{tabular}
\label{tab:quan_exp1}
\end{table}

The proposed method outperforms the baseline architecture regardless of the use of post-processing.
The benefit of using the \textbf{4P} module is clear when the disparity maps are obtained by using the pixel-wise winner-takes-it-all (WTA) rule without any post-processing.
Given that the images in the dataset contain many weakly-textured areas,
the small-sized $11\times11$ window cannot distinguish the true matches from false ones without the aid of post-processing.
On the other hand, the proposed architecture effectively sees the larger window, $37\times37$, by inserting the \textbf{4P} module before the final decision layer.

\begin{figure*}[t]
\centering
	\includegraphics[width=0.89\textwidth]{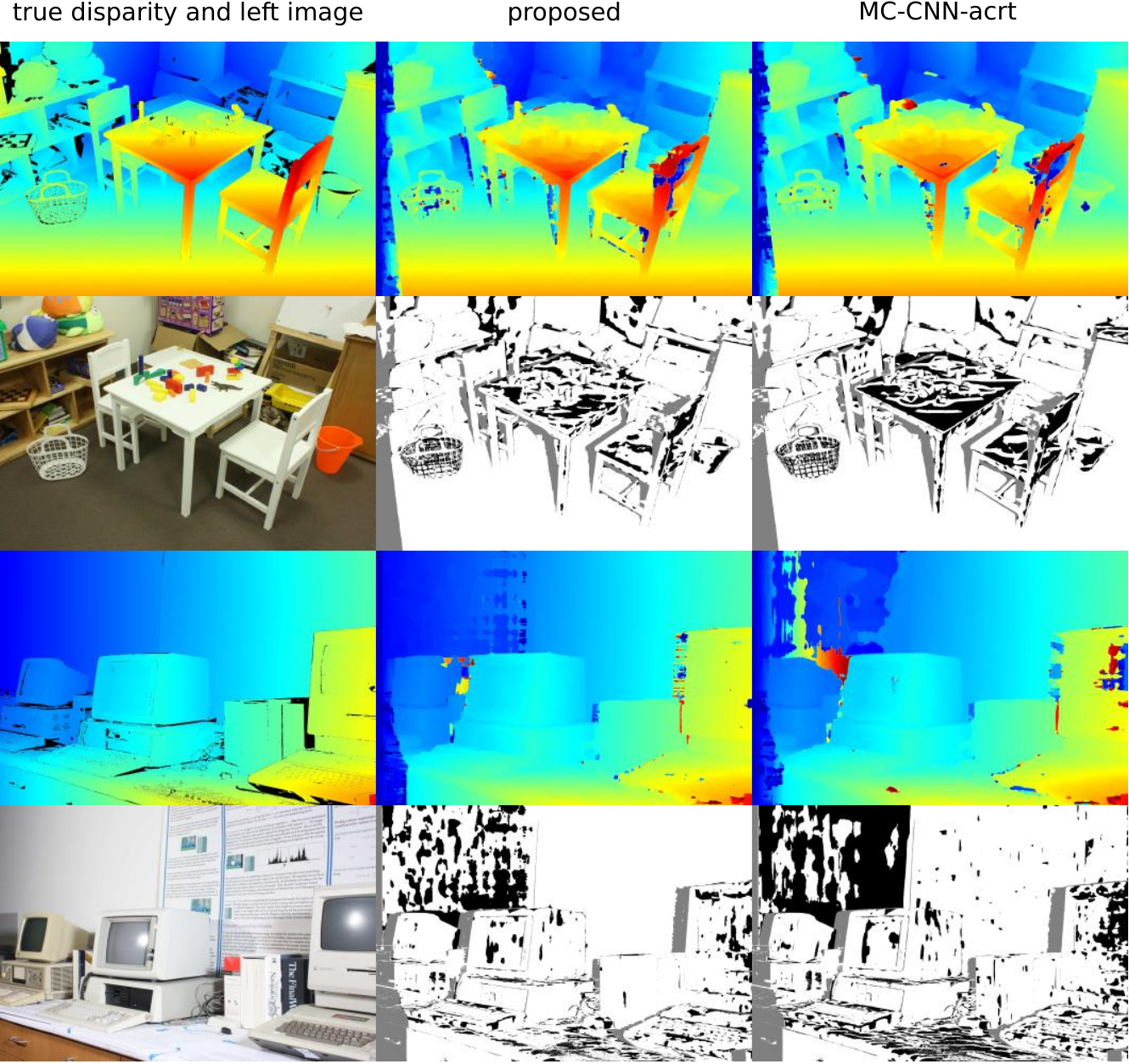}
	\caption{The results for \textbf{PlaytableP} and \textbf{Vintage} are visualized. For each datum, the upper row shows the disparity map and the bottom row shows the corresponding error maps. While the \lq{}\textbf{MC-CNN-acrt}\rq{}~\cite{vzbontar2016stereo} shows errors around the weakly-textured areas, such as the surfaces of the chair and the table in \textbf{PlaytableP} or the white wall in  \textbf{Vintage}, the proposed method shows more reliable results. }
\label{fig:qual_exp2}
\end{figure*}

It is less straightforward to understand why the proposed architecture still outperforms the baseline even after post-processing.
In that sense, it is worth to mention that the best parameter setting for post-processing of the proposed method largely differ from that of the baseline.\footnote{%
Following the conventions in~\cite{vzbontar2016stereo}, the best parameter setting is as follows: \texttt{cbca\_num\_iterations\_1 = 0},  \texttt{cbca\_num\_iterations\_2 = 1}, \texttt{sgm\_P1 = 1.3}, \texttt{sgm\_P2 = 17.0}, \texttt{sgm\_Q1 = 3.6}, \texttt{sgm\_Q2 = 36.0}, and \texttt{sgm\_V = 1.4}.}
The most notable changes from the original parameter setting is that we use much less number of CBCA~\cite{zhang2009cross}, and it means that multiple uses of CBCA~\cite{zhang2009cross} become redundant in the proposed architecture.
From this fact, we can interpret the role of \textbf{4P} module as adaptive local feature aggregation.
Compared to the hand-designed algorithm such as CBCA~\cite{zhang2009cross}, the influence of neighboring pixels to a certain pixel is automatically learned and it can be jointly trained with the cost function itself.
Furthermore, the information exchange among pixels is done in feature space which contains richer contextual information than the final cost volume space.

Note that the improvement over the baseline clearly results neither from the use of extra layers nor from the use of more parameters, as the authors of~\cite{vzbontar2016stereo} already have shown that the additional use of fully-connected (FC) layers is less significant. Using two additional FC layers leads to an improvement of approximately $1.90\%$, whereas using the \textbf{4P} module results in a $21.42\%$ improvement in terms of average error.

The main contribution of the proposed method lies in learning a less ambiguous matching cost function by inspecting a larger area. 
Figure~\ref{fig:qual_exp2} shows that the proposed network actually works better around the weakly-textured area than the \lq{}\textbf{MC-CNN-acrt}\rq{}~\cite{vzbontar2016stereo}.
The quantitative and qualitative results of each dataset, including the ones in the \lq{}test dense\rq{} set, are available at the Middlebury benchmark~\cite{scharstein2002taxonomy} website.




\section{Conclusions}
Viewing a large area to estimate the dense pixel correspondence is necessary to fully utilize the texture information to achieve less ambiguous and more accurate matching. A conventional matching cost function fails because neighboring pixels on the same surface as the target pixel are unknown. In this paper, a novel CNN module is proposed to make the CNN structure handle a large image patch without losing the small details, which enables it to learn an intelligent matching cost function for large-sized windows. The learned cost function can discriminate the false matches for weakly-textured areas or repeating textures, and can also conserve the disparity discontinuities well/. The learned cost function achieves competitive performance on the Middlebury benchmark.


%


  \newpage
\ifCLASSOPTIONcaptionsoff
  \newpage
\fi



%
\bibliographystyle{IEEEtran}

\end{document}